\documentclass{article}
\usepackage{ijcai24}

\usepackage{times}
\usepackage{soul}
\usepackage{url}
\usepackage[hidelinks]{hyperref}
\usepackage[utf8]{inputenc}
\usepackage[small]{caption}
\usepackage{graphicx}
\usepackage{xcolor}
\usepackage{amsmath}
\usepackage{amsthm}
\usepackage{booktabs}
\usepackage{algorithm}
\usepackage{algorithmic}
\usepackage{float}
\usepackage[switch]{lineno}

\usepackage{multirow}
\usepackage{adjustbox}
\usepackage{mathtools}
\usepackage{subcaption}
\usepackage{svg}

\title{A Neural Column Generation Approach to the Vehicle Routing Problem with Two-Dimensional Loading and Last-In-First-Out Constraints}

\author{
Yifan Xia$^1$
\and
Xiangyi Zhang$^2$\thanks{Corresponding author.}
\affiliations
$^1$State Key Laboratory for Novel Software Technology, Nanjing University\\
$^2$1QB Information Technologies, Inc.\\
\emails
\href{mailto:yfxia@smail.nju.edu.cn}{yfxia@smail.nju.edu.cn}, 
\href{mailto:xyzhangarccos0@gmail.com}{xyzhangarccos0@gmail.com}
}

\begin{document}

\maketitle

\begin{abstract}
    The vehicle routing problem with two-dimensional loading constraints (2L-CVRP) and the last-in-first-out (LIFO) rule presents significant practical and algorithmic challenges. While numerous heuristic approaches have been proposed to address its complexity, stemming from two NP-hard problems: the vehicle routing problem (VRP) and the two-dimensional bin packing problem (2D-BPP), less attention has been paid to developing exact algorithms. Bridging this gap, this article presents an exact algorithm that integrates advanced machine learning techniques, specifically a novel combination of attention and recurrence mechanisms. This integration accelerates the state-of-the-art exact algorithm by a median of 29.79\% across various problem instances. Moreover, the proposed algorithm successfully resolves an open instance in the standard test-bed, demonstrating significant improvements brought about by the incorporation of machine learning models. Code is available at \url{https://github.com/xyfffff/NCG-for-2L-CVRP}.
\end{abstract}

\section{Introduction}
\begin{figure*}[tb]
    \centering
    \begin{subfigure}[b]{0.45\textwidth}
        \includegraphics[width=\textwidth]{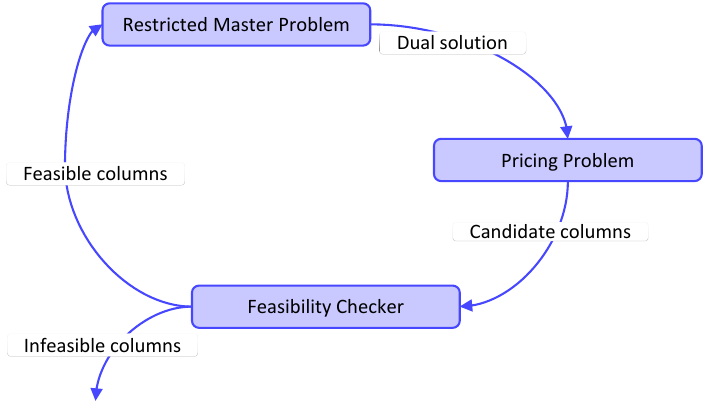}
        \caption{Pipeline of the state-of-the-art method.}
        \label{fig:sota_pipeline}
    \end{subfigure}
    \hfill 
    \begin{subfigure}[b]{0.45\textwidth}
        \includegraphics[width=\textwidth]{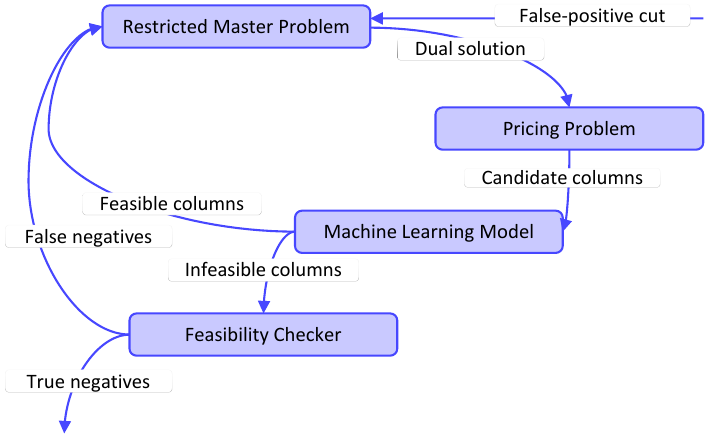}
        \caption{Pipeline of our neural column generation method.}
        \label{fig:ncg_pipeline}
    \end{subfigure}
    \caption{Comparative illustration of the SOTA method and our NCG method for the 2L-CVRP with the LIFO rule.}
    \label{fig:ncg}
\end{figure*}

The vehicle routing problem (VRP) is crucial in various logistics applications such as express systems, industrial warehousing, and on-demand delivery \cite{zong2022deep}. The basic form of VRP, or the capacitated vehicle routing problem (CVRP), focuses on optimizing a fleet of vehicles to meet customer demands with the goal of minimizing the total travel costs. An important extension of CVRP is the vehicle routing problem with the two-dimensional loading constraints (2L-CVRP) \cite{Iori_Salazar-González_Vigo_2007}. In 2L-CVRP, the challenge involves transporting rectangular items, each with specific length, width, and weight. The loading constraints in 2L-CVRP revolve around two main aspects: the orientation of items, which concerns whether an item can be rotated, and sequential loading, which concerns whether items belonging to successive customers along a route are allowed to be moved \cite{Fuellerer_Doerner_Hartl_Iori_2009}. The latter is often referred to as the last-in-first-out (LIFO) or rear-loading constraints. This problem is especially relevant in scenarios like transporting furniture or large industrial equipment where stacking items is not feasible due to safety or operational concerns. A real-world example of 2L-CVRP's application is seen at \href{www.opein.com}{Opein}, a company distributing large equipment, highlighting the model's practical importance. The practical value of 2L-CVRP has motivated numerous studies including both exact algorithms and heuristics, and solving new practical variants \cite{Pollaris_Braekers_Caris_Janssens_Limbourg_2015}. In this paper, we focus on 2L-CVRP with fixed item orientation and LIFO rule.

The 2L-CVRP encompasses two NP-hard problems: the CVRP and the two-dimensional bin packing problem (2D-BPP) stemming from the loading constraints. Given this complexity, the 2L-CVRP is solved approximately in most of studies \cite{gendreau2008tabu,Wei_Zhang_Zhang_Leung_2018}. However, the development of exact solvers is crucial for a deeper understanding of the problem's structure and for evaluating the performance gap between approximate solutions and optimal solutions. To achieve this end, recent studies \cite{cote2020vehicle,zhang2022branchcut,zhang2022branch} have proposed several efficient exact algorithms and closed many open instances. The state-of-the-art (SOTA) exact algorithm for the 2L-CVRP \cite{zhang2022branch} relies on the column generation (CG), which features repeatedly solving a challenging pricing problem (PP) and thus is regarded as one of the main bottlenecks blocking us from solving more open instances.

Recently, the integration of machine learning (ML) with CG has shown promise in solving combinatorial optimization problems more efficiently, while still aiming for optimal solutions \cite{mouad-columnselection,zhang2022learning,Shen:CG,chi2022a,morabit2023machine,yuan2023reinforcementlearningbased}. ML's ability to learn from data and make probabilistic predictions offers a potential acceleration in the CG process, particularly in problems where traditional methods are either too slow or impractical. 

However, applying ML-based CG methods to the 2L-CVRP, especially under the LIFO rule, introduces distinct challenges. Existing ML-based CG algorithms often solve the PP and generate multiple columns by traditional methods, with ML typically applied as a post-processing tool for column selection \cite{mouad-columnselection,chi2022a,yuan2023reinforcementlearningbased}. Some studies have attempted to bypass solving PP altogether, using ML to directly generate columns \cite{Shen:CG}. However, the PP in the context of 2L-CVRP, which combines the elementary shortest path problem with resource constraints (ESPPRC) and the 2D-BPP with the LIFO rule, is non-trivial to be addressed through traditional methods or ML models alone. While \cite{morabit2023machine} proposed an ML approach to prune the ESPPRC pricing graph, applying this strategy to the PP of 2L-CVRP is difficult due to the added complexity of 2D-BPP. A related work to ours is \cite{zhang2022learning}, which tackled a variant of 2L-CVRP without the LIFO constraint, utilizing feed-forward networks (FF) to accelerate the feasibility checking of columns generated by the pricing algorithm. Our study extends this approach by incorporating the LIFO rule, adding further complexity. We propose a novel ML model that leverages attention mechanism \cite{vaswani2017attention} for homogeneous items within the same customer and recurrence mechanism \cite{cho2014gru} for heterogeneous items across different customers. This model is used to predict the feasibility of columns generated from the ESPPRC, ensuring compliance with the 2D-BPP constraints with the LIFO rule. Our method provides a more efficient alternative to the traditional, time-consuming feasibility checker as exemplified in \cite{zhang2022branch}, achieving a median acceleration of 29.79\% and successfully solving one challenging open instance for the first time. For a visual representation of our algorithmic pipeline, see Figure \ref{fig:ncg_pipeline}. 

In summary, the main contributions of this paper are:
\begin{itemize}
\item We propose a neural column generation (NCG) algorithm that combines the state-of-the-art column generation for the 2L-CVRP with the newly developed ML model.
\item Our novel ML model integrates attention and recurrence mechanisms, along with a symmetry-based data augmentation technique. It effectively tackles the 2D-BPP with the LIFO rule, achieving an overall accuracy of around 95\%.
\item The NCG algorithm, when tested on standard benchmark instances, demonstrates a significant reduction in runtime, with a median decrease of 29.79\% compared to the state-of-the-art column generation algorithm.
\item Additionally, the NCG algorithm is incorporated into the state-of-the-art branch-and-price-and-cut (BPC) algorithm, successfully solving an open instance for the first time.
\end{itemize}

\section{Related Work}

In this section, we review the existing methods for 2L-CVRP and recent advances in enhance CG with ML techniques.

\paragraph{Non-learning Methods for 2L-CVRP.} The vehicle routing problem with two-dimensional loading constraints has been an active research area due to its practical relevance in logistics. The seminal work by \cite{Iori_Salazar-González_Vigo_2007} introduced a branch-and-cut (B$\&$C) algorithm adapted for 2L-CVRP by introducing infeasible-path constraints. They also propose an exact packing algorithm as a subroutine for the B$\&$C algorithm. The method set a benchmark for the problem, optimally solving instances with up to 35 customers. Subsequent heuristic approaches, such as the tabu search heuristic by \cite{gendreau2008tabu} and the guided local search by \cite{Zachariadis_Tarantilis_Kiranoudis_2009}, improved solution times and led to several new best solutions. The most effective heuristic for the 2L-CVRP to date involves a simulated annealing heuristic coupled with a local search-based packing algorithm \cite{Wei_Zhang_Zhang_Leung_2018}, which has shown superior solution quality. The 2L-CVRP with unloading sequential constraints has been tackled by \cite{Pinto_Alves_Carvalho_2013} through a column-generation-based heuristic. This approach initially omits loading constraints and then applies a heuristic to construct feasible packing solutions. The same authors later developed a more refined branch-and-price (B\&P) algorithm \cite{Pinto_Alves_de_Carvalho_2016} that integrates a variable neighborhood search (VNS) algorithm \cite{Pinto_Alves_de_Carvalho_2015} into the pricing problem. Despite these advancements, exact methods still face challenges with larger instances. The B\&C algorithm by \cite{cote2020vehicle} incorporated advanced exact packing algorithms \cite{Côté_Gendreau_Potvin_2014}, achieving a significant breakthrough in benchmark instances. \cite{zhang2022branchcut} further enhanced this approach by introducing a heuristic for separating infeasible set inequalities at fractional nodes, closing several open instances and improving dual bounds. Later on,  \cite{zhang2022branch} propose an exact CG algorithm addressing the loading constraints with a novel data structure \textit{L-Trie}. The CG algorithm leading to a successful BPC algorithm which closed many open instances for the first time and remains to be the state-of-the-art exact column generaiton for the 2L-CVRP. For additional studies on 2L-CVRP, please refer to \cite{Wang2009ASO,Pollaris_Braekers_Caris_Janssens_Limbourg_2015,Iori_de_Lima_Martello_Miyazawa_Monaci_2021}.
\paragraph{Neural Column Generation.} Recent years researchers have become increasingly interested in ML to accelerate optimization tasks \cite{BENGIO2021405}, and several learning-based methods have been proposed for specific problems solved by CG. For example, \cite{mouad-columnselection} employed a Graph Neural Network (GNN) for column selection in CG as a supervised binary classification task, aiming to mitigate degeneracy in the restricted master problem (RMP). This approach involved representing columns and constraints as a bipartite graph, with the GNN predicting whether to include or exclude each column during CG iterations. Building on this, \cite{chi2022a,yuan2023reinforcementlearningbased} modeled CG as a Markov decision process, using reinforcement learning for column selection during CG iterations. These methods, using GNN as Q-function approximator, proved to be more efficient than traditional greedy policies in problems such as the cutting stock problem. All these methods could be viewed as a post-processing step after CG, i.e., applying ML models to select from the generated columns rather than selecting greedily. \cite{morabit2023machine} proposed a supervised ML-based algorithm to prune the pricing network in the form of ESPPRC, alternating between reduced and complete graphs to accelerate the CG process. \cite{Shen:CG} utilized a support vector machine (SVM) for the graph coloring problem, directly generating columns by sampling from the SVM to enhance the CG process. The most related work to ours is \cite{zhang2022learning}, which addressed the vehicle routing problem without the LIFO rule. They utilized FF for heuristic validation of columns generated from the ESPPRC, reducing the dependence on the exact solver \cite{Côté_Dell’Amico_Iori_2014}. However, our paper addresses a more complex variant of the problem by including additional loading constraints, specifically the LIFO rule, which significantly increases the complexity of the CG process.

\section{Background}

In this section, we first present the mathematical formulation of the 2L-CVRP with the LIFO rule as well as the CG process. Then, we introduce the SOTA exact method for solving the 2L-CVRP with the LIFO rule.

\subsection{Problem Formulation} \label{formulation}

The 2L-CVRP with the LIFO rule is defined on a complete undirected graph $G=(V,E)$, where $V = \{0,1,2,...,n,n+1\}$ stands for the set of vertices consisting of customers $V_c = \{1,2,...,n\}$ and the depot $0$. Vertex $n+1$ represents a copy of the depot. The connections between any pair of vertices are depicted by the edge set $E$. $\forall e\in E$, $c_e$ represents the traveling cost associated with edge $e$. An alternative representation of an edge $(v_i, v_j)$ is also used. Set $K$ represents a fleet of homogeneous vehicles which are available at the depot. A loading area characterized by $H$ and $W$ as the length and width, respectively, is a property attached to each vehicle. The loading areas of all the vehicles are the same. Naturally, we have the total area of the loading surface of any vehicle equal to $A = H \times W$. Each vehicle also has a weight capacity denoted as $Q$.

As for the customers, $\forall i \in V_c$ is characterized by a set $M_i$ of rectangles. Any item $m \in M_i$ is marked by width $w_{i,m}$, length $h_{i,m}$, and weight $q_{i,m}$. Let $\nu_i$ and $c_i$ represent the total area and the total weights of all the items in customer $i$. In other words, $\nu_i = \sum_{m\in M_i}w_{i,m}h_{i,m}$ and $q_i = \sum_{m\in M_i}q_{i,m}$. The total number of the items in $G$ is equal to $|M|$, where $M$ is the union of the item sets of all the customers. The 2L-CVRP calls for planning routes for the fleet such that the demands of the customers are covered while respecting the following constraints:
\begin{enumerate}
    \item Each customer has to be visited exactly once;
    \item The total weight of the items to be delivered by a vehicle cannot exceed $Q$;
    \item The carried items have to be packed in the loading area without collision;
    \item Items are not allowed to be rotated over the course of packing;
    \item Items delivered to subsequent customers cannot be moved when unloading the items for the current customer (the LIFO rule).
\end{enumerate}

The 2L-CVRP can be formulated as a set partitioning (SP) problem:
\begin{align}
    \text{min} \, & \sum_{\mathclap{r\in \Omega}}c_r\lambda_r, \label{cstr:sp_obj} \\
    \text{s.t.} \, & \sum_{\mathclap{r\in \Omega}}\lambda_r = |K|, \label{cstr:sp_fleetsize} \\
    & \sum_{\mathclap{r\in \Omega}}a_{i,r}\lambda_r = 1, \ \forall i\in V_{c}, \label{cstr:sp_vertex_cover} \\
    & \lambda_r \in \{0,1\}, \ \forall r\in \Omega, \label{cstr:sp_nature}
\end{align}
where $\Omega$ represents the collection of feasible routes; $c_r$ represents the total traveling cost of route $r$; $\lambda_r$ is a binary decision variable indicating whether route $r$ is selected as a part of the solution; $a_{i,r}$ is a binary indicator, where $a_{i,r} = 1$ if vertex $i$ is visited in route $r$, and $a_{i,r} = 0$ otherwise. 

Equation (\ref{cstr:sp_obj}) defines the objective function for the set partitioning problem. Constraint set (\ref{cstr:sp_fleetsize}) imposes that there should be exactly $|K|$ routes to be selected as we assume that there is no idle vehicle in the fleet. This has been a convention when it comes to developing exact algorithms for the 2L-CVRP \cite{Iori_Salazar-González_Vigo_2007}. Constraint set (\ref{cstr:sp_vertex_cover}) calls for that each customer should be visited exactly once. Constraint set (\ref{cstr:sp_nature}) defines the domain of the decision variables.

In practical applications, directly solving the SP formulation is infeasible due to the need to enumerate all routes in $\Omega$. Typically, a smaller subset of $\Omega$ is selected to create a reduced version of the problem, known as the restricted master problem. Solving the RMP yields a solution, $\lambda^*$, that minimizes the objective value of the reduced formulation. However, $\lambda^*$ may not be optimal for the original problem. To potentially improve upon $\lambda^*$, a sub-problem called the pricing problem is solved to identify any routes (or columns) in $\Omega$ that could enhance the solution. This iterative process, known as column generation, continues until no further improvements are found. For a detailed discussion on this approach, readers are referred to Chapter 2 in \cite{desaulniers2006column}.

The formulation of the underlying pricing problem is as follows:
\begin{align}
    \text{min} \, & \sum_{\mathclap{e\in E}}\bar{d}_ex_e - \pi_f, \label{cstr:pp_obj_func} \\
    \text{s.t.} \, & \sum_{\mathclap{e\in \delta(i)}}x_e = 2, \ \forall i\in V, \label{cstr:pp_degree} \\
    & \sum_{\mathclap{e\in \delta(S)}}x_e \geq 2,\ \forall S\subset V_c, 1<|S|<n-1, \forall i \in S, \label{cstr:pp_sec} \\
    & \sum_{\mathclap{(i,j)\in E}}x_{ij}(q_i + q_j) \leq 2Q, \label{cstr:pp_capacity} \\
    & \sum_{\mathclap{(i,j) \in E}} x_{ij}(\nu_i + \nu_j) \leq 2A, \label{cstr:pp_area} \\
    & \sum_{\mathclap{e \in E(S, \sigma)}}x_e \leq |S| - 1, \ \forall (S, \sigma)\text{ such that }\sigma \notin \Sigma(S), \label{cstr:pp_infeasible_path} \\
    & x_e\in\{0,1\},\  \forall e\in E, \label{cstr:pp_integerality}
\end{align}
where $\bar{d}_e$ is the reduced cost defined as $\bar{d}_{i,j} = c_{i,j} - \frac{1}{2} \pi_i - \frac{1}{2} \pi_j, \ \forall (i,j)\in E$. The binary decision variable $x_e$ indicates whether edge $e$ in $E$ is used. The set $\delta(i)$ denotes edges incident to vertex $i$, and $\delta(S)$ denotes edges connecting vertices inside $S$ with those outside, where $S$ is a subset of $V$. $\Sigma(S)$ is the set of all feasible permutations of vertices in $S$. The route constructed by set $S$ in order $\sigma$ is represented as $(S, \sigma)$, with $E(S, \sigma)$ being its edge set. The dual variables $\pi_i$ and $\pi_f$ are associated with constraints \ref{cstr:sp_vertex_cover} and \ref{cstr:sp_fleetsize}, respectively.

The objective function (constraint set \ref{cstr:pp_obj_func}) aims to minimize the reduced cost. Constraint set (\ref{cstr:pp_degree}) ensures proper degree constraints for vertices, and constraint set (\ref{cstr:pp_sec}) addresses subtour elimination \cite{desrochers1991improvements}. Constraint sets (\ref{cstr:pp_capacity}) and (\ref{cstr:pp_area}) ensure that the vehicle's capacity and loading surface area limits are not exceeded. Finally, constraint set (\ref{cstr:pp_infeasible_path}) imposes restrictions due to the loading constraints.




\subsection{State-of-the-Art Pipeline}

Figure \ref{fig:sota_pipeline} illustrates the pipeline of the SOTA CG-based algorithm for the 2L-CVRP with the LIFO rule, as proposed by \cite{zhang2022branch}. The process starts with a restricted master problem which essentially enumerates a subset of set $\Omega$. The next step involves solving the linear relaxation of this problem to obtain a dual solution, which then facilitates the establishment of the PP. The labeling algorithm, enhanced by trie \cite{Brass_2010}, completion bounds, and ng-route relaxation \cite{Baldacci_Mingozzi_Roberti_2011}, efficiently prices out improving columns without checking the loading feasibility. The last step involves filtering out infeasible columns with the exact checker \cite{Côté_Gendreau_Potvin_2014} and adding the feasible ones to the restricted master problem, iterating until no feasible column is found. For a more detailed understanding, please refer to \cite{zhang2022branch}.

\section{Methodology}

In this section, we first introduce the motivation behind our integrated NCG approach as well as the outline of our approach. Then, we elaborate on the architecture of our machine learning model, including two key mechanisms and an important data augmentation technique.

\subsection{Motivation and Outline}

The current SOTA pipeline employs an exact feasibility checker to evaluate each candidate column generated by the labeling algorithm, necessitating solving the 2D-BPP with LIFO rule, a known NP-hard problem. 
Our proposed approach, however, introduces a ML model to refine this process by reducing the dependency on solving 2D-BPP. As illustrated in Figure \ref{fig:ncg_pipeline}, the ML model is positioned before the feasibility checker, categorizing candidate columns into `feasible' and `infeasible'. Feasible columns directly enter the restricted master problem, while infeasible ones are further checked to rectify potential false negatives and thus preventing optimal columns being discarded. To manage false positives, the loading feasibility of variables in the optimal basis is checked during master problem resolution, adding false-positive cuts to exclude infeasible variables from the feasible solution space.

This approach leverages the insight that columns not part of the optimal basis can bypass exact checking. 
However, its success depends on the ML model's accuracy, since frequent erroneous predictions can increase computational iterations.

\subsection{Machine Learning Model} \label{Machine Learning Model}
\begin{figure*}[tb]
\centering
\includegraphics[width=\textwidth]{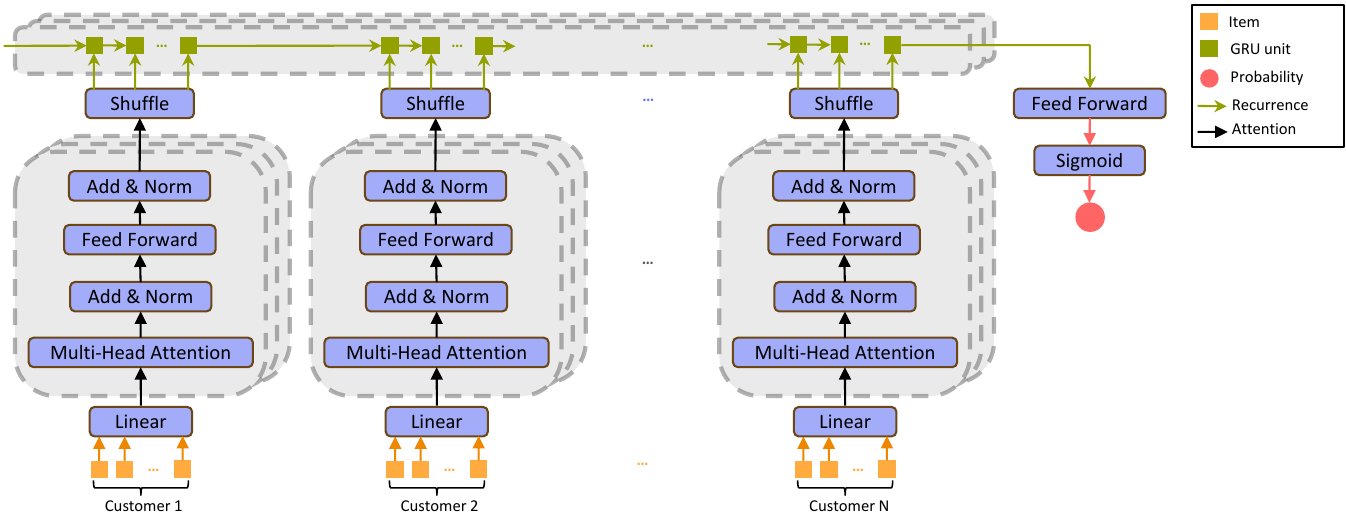}
\caption{The architecture of the machine learning model employed in our NCG approach, illustrating the integration of attention and GRU mechanisms for column feasibility prediction.}
\label{architechture}
\end{figure*}
Note that candidate columns generated by the labelling algorithm might violate the loading constraints. To address this, we frame it as a binary classification problem, developing a ML model to predict the feasibility of each candidate column. Specifically, our model predicts the probability with which each column respects the loading constraints.

Our model's architecture comprises a parallel embedding mechanism and a recursive processing strategy, designed to effectively capture both homogeneous and heterogeneous features of items within each column. For a given input candidate column, represented as a sequence of item sets $\left[ \left\{ x_{i,m} \right\}_{1 \leq m \leq \left|M_{i}\right|}\right]_{1 \leq i \leq n}$, each item $x_{i,m}$ is represented by normalized dimensions $\left[\frac{w_{i,m}}{W}, \frac{h_{i,m}}{H}\right]$. We employ an attention mechanism \cite{vaswani2017attention} for each customer $i$ to integrate features of homogeneous items (those belonging to the same customer). To handle heterogeneous features (items across different customers), a GRU \cite{cho2014gru} is utilized to process the sequence, incorporating the order information which is crucial in adhering to the LIFO constraints. Moreover, a symmetry-based data augmentation technique is employed to incorporate permutation invariance into the model. For a visual representation of our model's architecture, please refer to Figure \ref{architechture}.

\subsubsection{Item-Level Attention Mechanism} \label{item-level}

In the 2L-CVRP context, items belonging to the same customer are homogeneous and exempt from LIFO constraints, making the Multi-Head Attention (MHA) mechanism a suitable choice \cite{Kool2018AttentionLT,NEURIPS2020_pomo}. Our model uses MHA to capture the shared features of items within the same customer. The initial embedding for the $i$-th customer's $m$-th item is formulated as $h_{i,m}^{0} = W_{0}x_{i,m} + b_{0}$, where $W_{0}$ is the initial projection matrix and $b_{0}$ is the bias vector. The architecture includes skip-connections \cite{He2015resnet}, feed-forward networks, and layer normalization (LN) \cite{ba2016layernorm} in each sublayer. The embedding for the $i$-th customer's $m$-th item is iteratively updated in the $l$-th layer, as depicted in the following equations:
\begin{equation}
    \hat{h}_{i,m}^{l} = \text{LN}^{l}\left(h_{i,m}^{l-1} + \text{MHA}_{i,m}^{l}\left(h_{i,1}^{l-1}, \ldots, h_{i,\left|M_{i}\right|}^{l-1}\right)\right),
\end{equation}
\begin{equation}
    h_{i,m}^{l} = \text{LN}^{l}\left(\hat{h}_{i,m}^{l} + \text{FF}^{l}\left(\hat{h}_{i,m}^{l}\right)\right).
\end{equation}

The MHA mechanism at the core of our attention layer is defined as follows:
\begin{equation}
    Q_{i,m}^{j}, K_{i,m}^{j}, V_{i,m}^{j} = W_{Q}^{j}h_{i,m}, W_{K}^{j}h_{i,m}, W_{V}^{j}h_{i,m},
\end{equation}
\begin{equation}
    A_{i,m}^{j} = \text{softmax}\left( Q_{i,m}^{j}{K^{j}}^{T} / \sqrt{d_{k}} \right)V^{j},
\end{equation}
\begin{equation}
    \text{MHA}_{i,m} = \text{Concat}\left(A_{i,m}^{1}, A_{i,m}^{2}, ..., A_{i,m}^{H}\right)W_{O},
\end{equation}

where $j = 1, 2, ..., H$ and $d_{k} = d_{h} / H$. Here, $H$ is the number of attention heads, $d_{h}$ is the dimension of the item embedding, and $Q_{i,m}^{j}, K_{i,m}^{j}, V_{i,m}^{j}$ represent the query, key, and value vectors, respectively. $W_{O}$ is the projection matrix utilized to project the final MHA output. The final embedding of each item after $L$ layers is denoted by $h_{i,m} = h_{i,m}^{L}$.

\subsubsection{Customer-Level Recurrence Mechanism}

In the context of the 2L-CVRP with the LIFO rule, a sequential order relationship exists among customers, indicating that items from different customers are inherently heterogeneous. This order relationship dictates that for customer $i$ visited before customer $i+1$, items of customer $i$ (denoted as $x_{i,1}, x_{i,2}, \ldots, x_{i,\left|M_{i}\right|}$) must be loaded into the vehicle after items of customer $i+1$ ($x_{i+1,1}, x_{i+1,2}, \ldots, x_{i+1,\left|M_{i+1}\right|}$).

To model the recursive relationships among items of different customers, we use GRU as follows:
\begin{equation}
    \widetilde{h}_{t} = \text{GRU}\left(h_{t}, \widetilde{h}_{t-1}\right),
\end{equation}
where $\widetilde{h}_{t}$ denotes the hidden state at time step $t$ and $\widetilde{h}_{0} = \mathbf{0}$. 
In our approach, customers are processed sequentially, inputting one item $h_t$ per time step $t$ into the GRU, with the total number of time steps $T$ equaling the total items, $T = \sum_{i=1}^{n}\left|M_i\right|$.
Upon processing the last item of the last customer, the final state $\widetilde{h}_{T}$ of the GRU is transformed into a probability via a FF network and a sigmoid function:
\begin{equation}
    \textit{probability} = \text{sigmoid}\left( \text{FF}\left( \widetilde{h}_{T} \right) \right).
\end{equation}

\subsubsection{Data Augmentation with Permutation Invariance} \label{permutation}

As discussed in Section \ref{item-level}, items belonging to the same customer are homogeneous and are not constrained by any specific order. This characteristic allows for the application of permutation invariance as a data augmentation strategy, reflecting the symmetry of combinatorial problems \cite{NEURIPS2020_pomo,kim2022symnco}. By permuting the items of each customer, we can generate new, equivalent instances, expanding the training dataset and mitigating early overfitting.

Specifically, after applying the item-level multi-head attention mechanism to all items of each customer $i$, we perform a permutation $\pi_{i}$, shuffling the sequence $(1, 2, ..., \left|M_{i}\right|)$ to produce varied item orderings. This process can be represented mathematically as:
\begin{equation}
h_{i,1}, h_{i,2}, ..., h_{i, \left|M_{i}\right|} = h_{i,\pi_{i}\left(1\right)}, h_{i,\pi_{i}\left(2\right)}, ..., h_{i, \pi_{i}\left(\left|M_{i}\right|\right)},
\end{equation}

This permutation is applied independently to each customer's set of items before they are processed by the GRU.

\section{Experiments} \label{Sec:experimental_results}
\begin{figure}[tb]
\centering
\includegraphics[width=\linewidth]{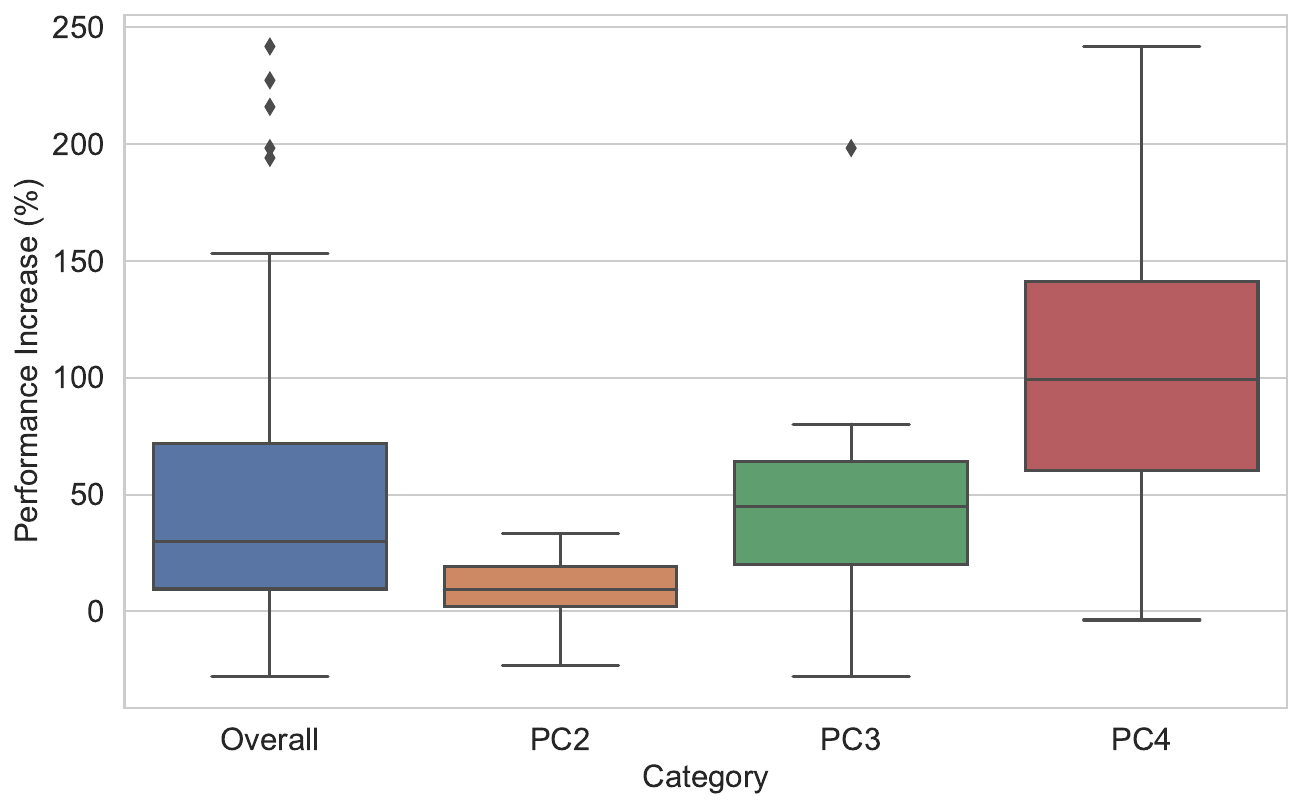}
\caption{Box plot depicting the percentage increase in efficiency of our NCG method over the SOTA across different instance categories.}
\label{fig:box_runtime}
\end{figure}
\subsection{Experimental Settings}

\paragraph{Data Sets.} Our machine learning model is trained on a unique set of packing problem instances, distinct from the benchmark instances used in \cite{Iori_Salazar-González_Vigo_2007}, ensuring an unbiased performance evaluation. The training dataset is derived by solving the 2L-CVRP problems as well as 2L-VRPTW (the 2L-CVRP with time window constraints). In particular, we stored all the packing instances in the course of solving the routing problems. There are in total two batches of the training samples: one from 2L-VRPTW instances and another from 2L-CVRP instances. The VRP instances are randomly generated as per the description in \cite{zhang2022learning}, ensuring a diverse distribution distinct from benchmark instances. The benchmark instances are categorized into five families, as defined in \cite{Iori_Salazar-González_Vigo_2007}. For details, please refer to the Supplementary Material. We exclude family 1, as it is equivalent to the one-dimensional loading scenario, and family 5, as highlighted by \cite{zhang2022branch}, due to its focus on very small items and a specialized column generation variant, making it less relevant for testing our algorithm in standard 2L-CVRP contexts. In Section \ref{Sec:experimental_results}, terms 'family' and 'packing class (PC)' are used interchangeably.

\paragraph{Evaluation Metrics.} For the 2L-CVRP instances, our experiment compares the NCG approach to the SOTA CG method by solving the linear relaxation of Formulation \ref{cstr:sp_obj} - \ref{cstr:sp_nature}. Given that both algorithms are exact, their optimal objective values are expected to match. Our primary interest lies in the difference in computational time. Let $T_{NCG}$ represent the walltime for NCG and $T_{SOTA}$ for SOTA CG. The percentage gap, calculated as $\frac{T_{SOTA} - T_{NCG}}{T_{NCG}} \times 100\%$, serves to quantify the time savings and performance increase, with higher values indicating greater efficiency gains for NCG. Furthermore, we track the number of column generation iterations to evaluate the impact of the ML-induced false-positive cuts on the frequency of column generation.

\paragraph{Hardware.} During the training phase of the ML model, experiments were conducted on an AMD EPYC 7V13 64-Core CPU @ 2.45GHz with an NVIDIA A100 GPU. Post training, the ML model was serialized and integrated into the NCG algorithm using Torch C++. To ensure a fair comparison, both the NCG algorithm and the SOTA algorithm \cite{zhang2022branch} were implemented in C++ and evaluated on an Intel i5-10600KF processor @ 4.10 GHz.

\subsection{Main Results}

\begin{figure}[tb]
\centering
\includegraphics[width=\linewidth]{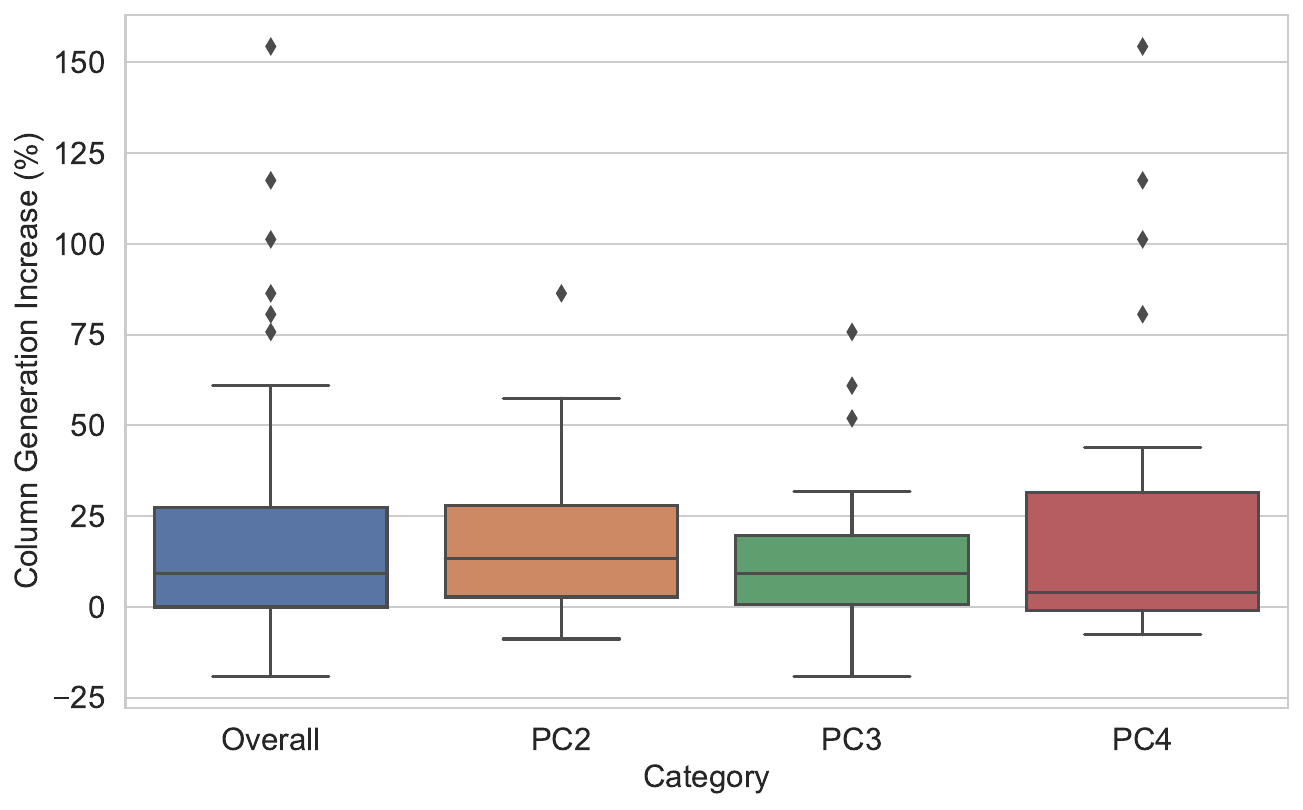}
\caption{Box plot demonstrating the percentage increase in column generation counts for our NCG method over the SOTA.}
\label{fig:box_cg}
\end{figure}

\begin{table*}[tb]
    \centering
    \begin{tabular}{cccccc}
    \toprule
\multirow{2}{*}{Instance Name} & \multicolumn{2}{c}{NCG-based BPC} & \multicolumn{3}{c}{SOTA BPC} \\ 
\cmidrule(lr){2-3}
\cmidrule(lr){4-6}
  & Optimal Obj & Run Time (s)  & Lower Bound & Primal Bound & Run Time (s)\\ 
  \midrule
2l-cvrp-2304    & 1068         & 49544.6 & 1066.78 & 1069 &  69948.96$^\dagger$  \\
    \bottomrule
\end{tabular}
    \\
    \hspace{-200pt}\footnotesize{$\dagger$ Algorithm terminated early due to memory overload.}
    \caption{Performance comparison on the solved open instance \textit{2304} between NCG-based BPC and SOTA BPC.}
    \label{tab:solved_instance}
\end{table*}

\begin{table*}[tb]
    \centering
    \begin{tabular}{lcccccccc}
    \toprule
    \multirow{2}{*}{Algorithm} & \multicolumn{2}{c}{Overall} & \multicolumn{2}{c}{PC2} & \multicolumn{2}{c}{PC3} & \multicolumn{2}{c}{PC4} \\
    \cmidrule(lr){2-3}
    \cmidrule(lr){4-5}
    \cmidrule(lr){6-7}
    \cmidrule(lr){8-9}
    & TPR & TNR & TPR & TNR & TPR & TNR & TPR & TNR \\
    \midrule
    Ours & \textbf{94.08\%} & \textbf{96.80\%} & \textbf{84.26\%} & \textbf{97.33\%} & \textbf{86.79\%} & 95.83\% & \textbf{97.81\%} & \textbf{88.81\%} \\
    w/o augmentation & 92.84\% & 96.78\% & 81.54\% & 97.22\% & 83.07\% & \textbf{96.69\%} & 97.47\% & 87.31\% \\
    w/o attention mechanism & 91.44\% & 95.22\% & 77.56\% & 95.88\% & 80.76\% & 94.68\% & 96.43\% & 82.84\% \\
    w/o recurrence mechanism & 92.48\% & 96.07\% & 81.70\% & 96.86\% & 81.65\% & 93.81\% & 95.40\% & 88.06\% \\
    \bottomrule
    \end{tabular}
    \caption{Comparative results showcasing TPR and TNR metrics across different model configurations for Overall and PC2, PC3, PC4 categories.}
    \label{tab:customized}
\end{table*}
\subsubsection{Acceleration of SOTA Algorithm}
We evaluated the NCG against the SOTA on benchmark instances with up to 50 customers from families 2 - 4. As depicted in Figure \ref{fig:box_runtime}, NCG demonstrates a significant advantage over the SOTA algorithm, with a median performance improvement of 29.79\% across all tested instances. Notably, the performance gains are more pronounced in PC3 and PC4, with median percentage gaps of 44.97\% and 99.22\%, respectively. This improvement correlates with the higher accuracy of our ML model in PC3 and PC4, as indicated in Table \ref{tab:customized}, suggesting fewer iterations required to rectify false predictions. In contrast, the least improvement occurs with PC2, with with a median gain of 9.38\%, as the ML model reaches the worst accuracy in PC2.

Figure \ref{fig:box_cg} also presents an intriguing observation: an increase in the number of CG iterations due to the incorporation of false positive cuts. Interestingly, for each instance family, the increase in CG iterations inversely correlates with performance improvement. Despite this, the additional time incurred by these extra iterations is more than compensated for by the time savings from the ML model. This highlights the effectiveness of our NCG approach in enhancing overall computational efficiency despite the potential for additional CG iterations. For detailed results on each instance, please refer to the Supplementary Material.

\subsubsection{Solving Open Instance}
Our NCG algorithm was integrated into the BPC framework developed in \cite{zhang2022branch}. This integrated approach was tested on benchmark instances, leading to the successful resolution of an open instance. As shown in Table \ref{tab:solved_instance}, for the instance \textit{2304}, our NCG-based BPC method obtained an optimal objective value (Optimal Obj) of 1068, while the existing SOTA BPC algorithm was unable to reach an optimal solution, yielding only a lower bound of 1066.78 and a primal bound of 1069, despite running for approximately 1.4 times longer than our approach. Furthermore, the SOTA BPC algorithm suffered from memory overload, whereas our NCG-based BPC approach did not, indicating its more efficient memory usage. This efficiency is attributed to the reduced number of routes that needed to be checked by the exact packing algorithm, resulting in lower memory demands for the BPC algorithm.

\subsection{Ablation Study}

\begin{figure}[tb]
\centering
\includegraphics[width=\linewidth]{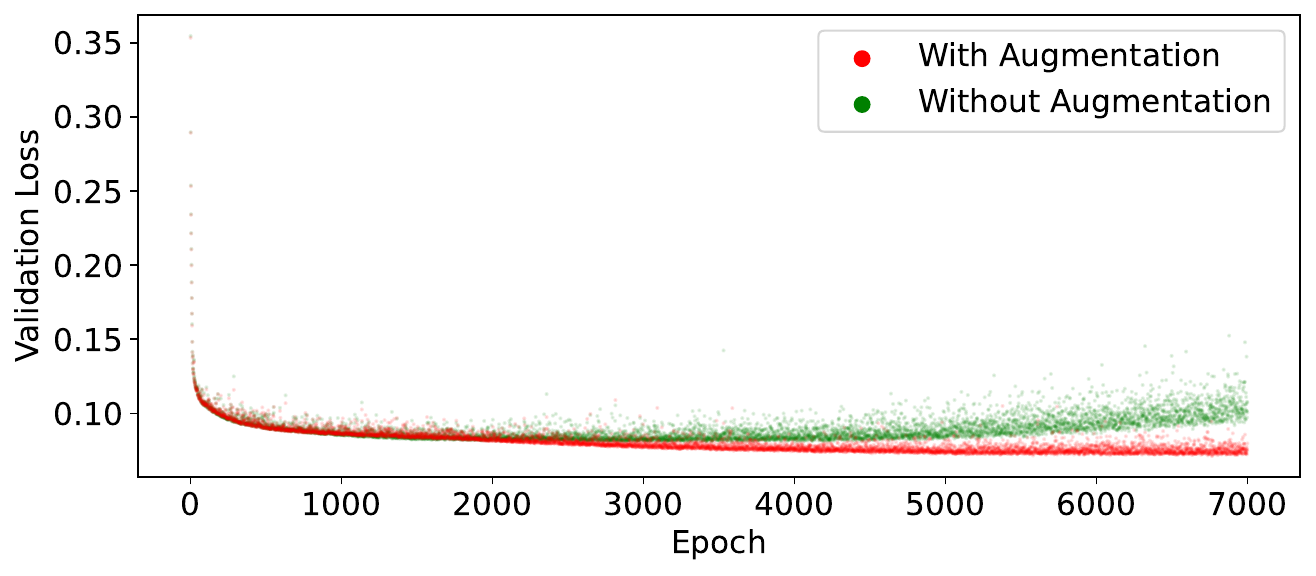}
\caption{Comparison of validation losses over epochs for the baseline and augmented models.}
\label{val_loss_comparison}
\end{figure}
To validate the contributions of the attention and recurrence mechanisms and the data augmentation technique described in Section \ref{Machine Learning Model}, we performed an ablation study. We compared the following configurations:
\begin{enumerate}
    \item The full NCG model that integrates both attention and recurrence mechanisms with data augmentation.
    \item A variant without data augmentation.
    \item A model where the attention mechanism is replaced by a multilayer perceptron (MLP).
    \item A model where the recurrence mechanism is replaced by a transformer encoder with sinusoidal positional encoding \cite{vaswani2017attention}.
\end{enumerate}

As shown in Table \ref{tab:customized}, the complete NCG model outperforms the other variants in terms of true positive rate (TPR) and true negative rate (TNR). These results suggest that the item-level attention mechanism contributes the most, followed by the customer-level recurrence mechanism, and then the data augmentation technique. Despite the transformer encoder's capability to process sequential information, it underperforms compared to the GRU-based model, highlighting the importance of explicit modeling of sequential relationships in the 2L-CVRP with the LIFO rule.

Figure \ref{val_loss_comparison} shows that, compared to the baseline model which overfits sooner, the model with permutation invariance maintains a lower validation loss for a longer period, confirming the effectiveness of the data augmentation strategy in enhancing generalization capacity.

\section{Conclusion}

In this work, we introduced a ML-based exact algorithm to solve the 2L-CVRP with the LIFO rule. Our approach integrates neural column generation into the SOTA pipeline, formulating the feasibility checking of candidate columns as a binary classification problem. To preserve solution optimality, we implemented post-processing steps for handling false negatives and positives. The ML model in our framework utilizes an attention mechanism for capturing homogeneous item features and a recurrence mechanism for heterogeneous features. Additionally, we employed data augmentation exploiting the problem's symmetry.

Experimental evaluations on benchmark instances demonstrate that our NCG method notably accelerates the SOTA algorithm by a median of 29.79\% across various problem instances. Significantly, our approach also successfully solves an open instance, marking a substantial contribution to the field. This achievement highlights the potential of integrating machine learning techniques into traditional optimization problems for enhanced performance and efficiency.



\bibliographystyle{named}
\bibliography{ijcai24}

\newpage
\appendix
\onecolumn

\section{Data Sets}

\subsection{Details of Training Samples}

To generate training samples for the machine learning model, we first created a set of 2L-VRPTW and 2L-CVRP instances. The reason we went for 2L-VRPTW because it provides us with more differences from the benchmark instances. The 2L-VRPTW were generated as per the following characteristics.
\begin{enumerate}
    \item \textbf{Customer distribution:} This is the distribution from which customer coordinates in $G=(V,E)$ are sampled. We designed three different scenarios: \textit{pure random}, \textit{clustered}, and \textit{mixed}. Specifically, for \textit{pure random}, the depot is fixed at (35,35) while all customer coordinates are uniformly distributed in the range [0,100]. In \textit{clustered}, the depot is at (40,50), each cluster center is sampled from [10,90], and the number of customers per cluster ranges from 8 to 9. Customers within a cluster are randomly placed at a distance from the cluster center, determined by a uniform distribution in the range [3,5]. For \textit{mixed}, 50\% of the customers are generated randomly, and the rest are clustered.
    \item \textbf{Time windows:} Generating time windows involves two steps. First, we determine the distribution for the time window "diameter" (half of the width). The mean is a fraction of the maximum return time to the depot, and the standard deviation is a product of the maximum return time and a value sampled from [0, 0.05, 0.10, 0.15, 0.20, 0.25]. The center of the time window is then uniformly sampled between the earliest and latest possible times to reach the customer. The time window is obtained by adding and subtracting the diameter from the center.
    \item \textbf{Rectangular items:} Items are generated based on Table \ref{tab:item_sizes}. The column \textit{PC} (Packing Class) specifies the packing class for each instance. The column $|M_i|$ indicates the number of items, where $M_i$ is the set of items for customer $i$. The master column \textit{Vertical} suggests that items should feature $h_{i,m} \ge w_{i,m}$ on average if $H = W$, \textit{Homogeneous} implies square items, and \textit{Horizontal} indicates items should feature $h_{i,m} \le w_{i,m}$ generally.
\end{enumerate}
The 2L-CVRP instances for training were generated similarly, except that time windows were not needed for customers. To generate the packing problems, we used the branch-and-price algorithm from \cite{zhang2022learning} with the ML model disabled for solving 2L-VRPTW, and the branch-and-price-and-cut algorithm from \cite{zhang2022branch} for 2L-CVRP.

\begin{table}[ht]
\centering
\begin{adjustbox}{width=\textwidth}
\begin{tabular}{lc|cc|cc|cc}
\toprule
\multirow{2}{*}{PC} & \multirow{2}{*}{$|M_i|$} & \multicolumn{2}{c|}{Vertical} & \multicolumn{2}{c|}{Homogeneous} & \multicolumn{2}{c}{Horizontal} \\
 &  & $h_{i,m}$ & $w_{i,m}$ & $h_{i,m}$ & $w_{i,m}$ & $h_{i,m}$ & $w_{i,m}$ \\
\midrule
2 & [1,2] & [4H/10, 9H/10] & [W/10, 2W/10] & [2H/10, 5W/10] & [2W/10, 5W/10] & [H/10, 2H/10] & [4W/10, 9W/10] \\
3 & [1,3] & [3H/10, 8H/10] & [W/10, 2W/10] & [2H/10, 4H/10] & [2W/10, 4W/10] & [H/10, 2H/10] & [3W/10, 8W/10] \\
4 & [1,4] & [2H/10, 7H/10] & [W/10, 2W/10] & [3H/10, 4H/10] & [W/10, 4W/10] & [H/10, 2H/10] & [2W/10, 7W/10] \\
5 & [1,5] & [H/10, 6H/10] & [W/10, 2W/10] & [H/10, 3H/10] & [W/10, 3W/10] & [H/10, 2H/10] & [W/10, 6W/10] \\
\bottomrule
\end{tabular}
\end{adjustbox}
\caption{Description of item sizes for each instance family}
\label{tab:item_sizes}
\end{table}

\subsection{Details of Benchmark Instances}
The benchmark instances were generated by \cite{Iori_Salazar-González_Vigo_2007} following Table \ref{tab:item_sizes} for item numbers and sizes. For customer coordinates, they used existing CVRP instances, generating one instance for each packing class from each CVRP instance. The number of vehicles for each instance was determined by solving a 2D-BPP problem to find the minimum number of bins needed to pack all items in $G$. The final vehicle count is the maximum of the 2D-BPP solution and the number of vehicles in the original CVRP instance.

\section{Experimental Results}
\begin{table}[H]
    \centering
    \small
    \begin{tabular}{lrrrr}
        \toprule
        Instance &  SOTA Run Time (s) &  NCG Run Time (s) &  Performance Increase (\%) &  PC \\
        \midrule
2l-cvrp0102-90\&3 & 18.61 & 18.18 & 2.38 & 2 \\
2l-cvrp1902-160\&11 & 79.69 & 72.73 & 9.57 & 2 \\
2l-cvrp1802-2010\&9 & 209.80 & 272.98 & -23.14 & 2 \\
2l-cvrp1702-60\&14 & 8.03 & 7.76 & 3.43 & 2 \\
2l-cvrp1602-67\&11 & 4.52 & 3.82 & 18.11 & 2 \\
2l-cvrp1402-8000\&7 & 70.26 & 77.25 & -9.05 & 2 \\
2l-cvrp1302-38000\&7 & 22.38 & 21.98 & 1.82 & 2 \\
2l-cvrp1202-68\&9 & 27.87 & 26.87 & 3.73 & 2 \\
2l-cvrp1102-4500\&6 & 61.24 & 55.78 & 9.79 & 2 \\
2l-cvrp1002-4500\&6 & 36.84 & 27.62 & 33.35 & 2 \\
2l-cvrp0902-48\&8 & 2.61 & 2.09 & 25.06 & 2 \\
2l-cvrp0802-4500\&5 & 9.70 & 10.00 & -2.99 & 2 \\
2l-cvrp1502-8000\&6 & 135.67 & 146.51 & -7.40 & 2 \\
2l-cvrp0202-55\&5 & 0.64 & 0.58 & 9.98 & 2 \\
2l-cvrp0602-4000\&6 & 4.65 & 4.47 & 4.16 & 2 \\
2l-cvrp0702-4500\&5 & 14.45 & 13.21 & 9.38 & 2 \\
2l-cvrp0502-6000\&4 & 11.18 & 8.69 & 28.65 & 2 \\
2l-cvrp0302-85\&5 & 3.49 & 2.69 & 29.79 & 2 \\
2l-cvrp0402-58\&6 & 2.75 & 2.28 & 20.34 & 2 \\
2l-cvrp1903-160\&11 & 123.66 & 106.06 & 16.60 & 3 \\
2l-cvrp0103-90\&3 & 45.78 & 15.34 & 198.34 & 3 \\
2l-cvrp1803-2010\&10 & 80.36 & 87.62 & -8.28 & 3 \\
2l-cvrp1703-60\&14 & 6.56 & 9.09 & -27.87 & 3 \\
2l-cvrp0203-55\&5 & 1.42 & 1.10 & 29.25 & 3 \\
2l-cvrp1603-67\&11 & 6.07 & 3.68 & 64.95 & 3 \\
2l-cvrp1503-8000\&6 & 344.18 & 363.51 & -5.32 & 3 \\
2l-cvrp1403-8000\&7 & 80.15 & 62.15 & 28.95 & 3 \\
2l-cvrp0703-4500\&5 & 15.73 & 9.64 & 63.13 & 3 \\
2l-cvrp0603-4000\&6 & 4.90 & 3.25 & 50.70 & 3 \\
2l-cvrp1303-38000\&7 & 33.65 & 23.15 & 45.36 & 3 \\
2l-cvrp1203-68\&9 & 10.49 & 9.20 & 14.04 & 3 \\
2l-cvrp0403-58\&6 & 4.16 & 2.97 & 40.16 & 3 \\
2l-cvrp1103-4500\&7 & 29.95 & 24.22 & 23.69 & 3 \\
2l-cvrp1003-4500\&6 & 47.06 & 31.31 & 50.29 & 3 \\
2l-cvrp0503-6000\&4 & 17.40 & 9.66 & 80.10 & 3 \\
2l-cvrp0903-48\&8 & 8.44 & 5.82 & 44.97 & 3 \\
2l-cvrp0803-4500\&5 & 15.96 & 9.64 & 65.66 & 3 \\
2l-cvrp0303-85\&5 & 8.78 & 5.29 & 65.87 & 3 \\
2l-cvrp1204-68\&9 & 23.39 & 12.44 & 88.05 & 4 \\
2l-cvrp1304-38000\&7 & 151.96 & 88.45 & 71.79 & 4 \\
2l-cvrp1404-8000\&7 & 309.57 & 136.66 & 126.53 & 4 \\
2l-cvrp1804-2010\&10 & 184.45 & 191.46 & -3.66 & 4 \\
2l-cvrp1604-67\&11 & 24.70 & 12.40 & 99.22 & 4 \\
2l-cvrp1704-60\&14 & 10.65 & 8.24 & 29.17 & 4 \\
2l-cvrp1104-4500\&7 & 141.24 & 127.72 & 10.59 & 4 \\
2l-cvrp1504-8000\&8 & 218.22 & 162.76 & 34.07 & 4 \\
2l-cvrp1004-4500\&7 & 85.22 & 33.68 & 153.04 & 4 \\
2l-cvrp0504-6000\&4 & 34.82 & 11.02 & 216.00 & 4 \\
2l-cvrp0804-4500\&5 & 32.42 & 9.49 & 241.83 & 4 \\
2l-cvrp0704-4500\&5 & 41.94 & 18.28 & 129.38 & 4 \\
2l-cvrp0604-4000\&6 & 17.48 & 8.56 & 104.22 & 4 \\
2l-cvrp0404-58\&6 & 8.32 & 5.15 & 61.80 & 4 \\
2l-cvrp0304-85\&5 & 5.38 & 3.39 & 58.64 & 4 \\
2l-cvrp0204-55\&5 & 3.32 & 1.01 & 227.34 & 4 \\
2l-cvrp0104-90\&4 & 7.94 & 2.70 & 194.15 & 4 \\
2l-cvrp0904-48\&8 & 9.33 & 5.08 & 83.80 & 4 \\
2l-cvrp1904-160\&12 & 142.75 & 69.12 & 106.53 & 4 \\
        \bottomrule
    \end{tabular}
    \caption{Comparison of Runtime Speedup between SOTA and NCG Algorithms}
    \label{tab:speedup-comparison}
\end{table}

\begin{table}[H]
    \centering
    \begin{tabular}{lrrrr}
        \toprule
        Instance &  SOTA CG Iteration &  NCG Iteration &  Iteration Increase (\%) &  PC \\
        \midrule
2l-cvrp0102-90\&3 & 46 & 47 & 2.17 & 2 \\
2l-cvrp1902-160\&11 & 117 & 173 & 47.86 & 2 \\
2l-cvrp1802-2010\&9 & 132 & 246 & 86.36 & 2 \\
2l-cvrp1702-60\&14 & 69 & 66 & -4.35 & 2 \\
2l-cvrp1602-67\&11 & 56 & 63 & 12.50 & 2 \\
2l-cvrp1402-8000\&7 & 80 & 126 & 57.50 & 2 \\
2l-cvrp1302-38000\&7 & 65 & 92 & 41.54 & 2 \\
2l-cvrp1202-68\&9 & 68 & 77 & 13.24 & 2 \\
2l-cvrp1102-4500\&6 & 95 & 121 & 27.37 & 2 \\
2l-cvrp1002-4500\&6 & 71 & 85 & 19.72 & 2 \\
2l-cvrp0902-48\&8 & 34 & 31 & -8.82 & 2 \\
2l-cvrp0802-4500\&5 & 53 & 63 & 18.87 & 2 \\
2l-cvrp1502-8000\&6 & 136 & 175 & 28.68 & 2 \\
2l-cvrp0202-55\&5 & 24 & 25 & 4.17 & 2 \\
2l-cvrp0602-4000\&6 & 42 & 45 & 7.14 & 2 \\
2l-cvrp0702-4500\&5 & 69 & 66 & -4.35 & 2 \\
2l-cvrp0502-6000\&4 & 61 & 58 & -4.92 & 2 \\
2l-cvrp0302-85\&5 & 42 & 48 & 14.29 & 2 \\
2l-cvrp0402-58\&6 & 30 & 31 & 3.33 & 2 \\
2l-cvrp1903-160\&11 & 123 & 162 & 31.71 & 3 \\
2l-cvrp0103-90\&3 & 47 & 38 & -19.15 & 3 \\
2l-cvrp1803-2010\&10 & 64 & 103 & 60.94 & 3 \\
2l-cvrp1703-60\&14 & 54 & 56 & 3.70 & 3 \\
2l-cvrp0203-55\&5 & 25 & 30 & 20 & 3 \\
2l-cvrp1603-67\&11 & 59 & 62 & 5.08 & 3 \\
2l-cvrp1503-8000\&6 & 136 & 239 & 75.74 & 3 \\
2l-cvrp1403-8000\&7 & 70 & 70 & 0 & 3 \\
2l-cvrp0703-4500\&5 & 61 & 63 & 3.28 & 3 \\
2l-cvrp0603-4000\&6 & 39 & 46 & 17.95 & 3 \\
2l-cvrp1303-38000\&7 & 52 & 62 & 19.23 & 3 \\
2l-cvrp1203-68\&9 & 58 & 57 & -1.72 & 3 \\
2l-cvrp0403-58\&6 & 35 & 35 & 0 & 3 \\
2l-cvrp1103-4500\&7 & 52 & 79 & 51.92 & 3 \\
2l-cvrp1003-4500\&6 & 58 & 68 & 17.24 & 3 \\
2l-cvrp0503-6000\&4 & 60 & 53 & -11.67 & 3 \\
2l-cvrp0903-48\&8 & 54 & 59 & 9.26 & 3 \\
2l-cvrp0803-4500\&5 & 70 & 71 & 1.43 & 3 \\
2l-cvrp0303-85\&5 & 42 & 48 & 14.29 & 3 \\
2l-cvrp1204-68\&9 & 73 & 75 & 2.74 & 4 \\
2l-cvrp1304-38000\&7 & 67 & 121 & 80.60 & 4 \\
2l-cvrp1404-8000\&7 & 73 & 71 & -2.74 & 4 \\
2l-cvrp1804-2010\&10 & 103 & 224 & 117.48 & 4 \\
2l-cvrp1604-67\&11 & 68 & 81 & 19.12 & 4 \\
2l-cvrp1704-60\&14 & 104 & 108 & 3.85 & 4 \\
2l-cvrp1104-4500\&7 & 82 & 165 & 101.22 & 4 \\
2l-cvrp1504-8000\&8 & 46 & 117 & 154.35 & 4 \\
2l-cvrp1004-4500\&7 & 52 & 48 & -7.69 & 4 \\
2l-cvrp0504-6000\&4 & 72 & 74 & 2.78 & 4 \\
2l-cvrp0804-4500\&5 & 47 & 55 & 17.02 & 4 \\
2l-cvrp0704-4500\&5 & 77 & 80 & 3.90 & 4 \\
2l-cvrp0604-4000\&6 & 44 & 46 & 4.55 & 4 \\
2l-cvrp0404-58\&6 & 30 & 34 & 13.33 & 4 \\
2l-cvrp0304-85\&5 & 30 & 29 & -3.33 & 4 \\
2l-cvrp0204-55\&5 & 24 & 24 & 0 & 4 \\
2l-cvrp0104-90\&4 & 22 & 21 & -4.55 & 4 \\
2l-cvrp0904-48\&8 & 58 & 57 & -1.72 & 4 \\
2l-cvrp1904-160\&12 & 89 & 128 & 43.82 & 4 \\
        \bottomrule
    \end{tabular}
    \caption{Comparison of CG iteration increase between SOTA and NCG Algorithms}
    \label{tab:cg-increase-comparison}
\end{table}



\newpage
\section{Model Details}

\subsection{Training Details of the ML Model}
The table below (Table \ref{tab:training_details}) outlines the key parameters and their respective values used during the training process.

\begin{table}[htb]
\centering
\begin{tabular}{lc}
\toprule
Parameter & Value \\
\midrule
Hidden Dimension & 16 \\
Number of Heads in MHA & 4 \\
Number of Layers in MHA & 2 \\
Number of GRU Layers & 1 \\
Batch Size & 512 \\
Learning Rate of Adam & 0.0001 \\
Data Augmentation Multiplicity & 10 \\
\bottomrule
\end{tabular}
\caption{Training Details of the ML Model}
\label{tab:training_details}
\end{table}

\subsection{Loss Function}
Our machine learning model employs a weighted binary cross-entropy loss function, tailored for class imbalance. The loss is computed as follows:
\begin{equation}
    \text{Loss}(p, y) = - \frac{1}{N} \sum_{i=1}^{N} \left[ w_{i} \cdot y_{i} \cdot \log(p_{i}) + (1 - y_{i}) \cdot \log(1 - p_{i}) \right]
\end{equation}
where $p$ denotes the model's probability output, $y$ is the target label, and $N$ is the number of observations. The weight $w_{i}$ for each observation is determined as the ratio of the number of positive samples to negative samples in the dataset. This weighting strategy aims to mitigate the impact of class imbalance on the loss function.

\subsection{Ablation Study Details}
In our ablation study, we evaluate the impact of various components of our ML model by comparing it against three variants:

\paragraph{Without Data Augmentation} This model variant, labeled as `w/o augmentation', is identical to our primary model except for the data augmentation aspect. Here, we set the Data Augmentation Multiplicity parameter to 1, effectively disabling the permutation-based data augmentation strategy.

\paragraph{Without Attention Mechanism} In this variant, labeled as `w/o attention mechanism', the attention mechanism is replaced with a MLP. The MLP consists of two linear layers with a ReLU activation function in between. The first linear layer maps the 2-dimensional input (normalized width and length of items) to the hidden size, followed by the ReLU activation and another linear layer that maps back to the hidden size. The computation is as follows:
\begin{equation}
    \text{MLP}(x) = W_{2} \cdot \text{ReLU}\left( W_{1}\cdot x + b_{1} \right) + b_{2},
\end{equation}
where $x$ is the input vector, $W_1$, $W_2$ are weight matrices, and $b_1$, $b_2$ are bias vectors.

\paragraph{Without Recurrence Mechanism} The third variant, labeled as `w/o recurrence mechanism', replaces the GRU-based recurrence mechanism with a Transformer encoder layer equipped with sinusoidal positional encoding. The positional encoding is added to each item's representation to incorporate sequence information. After adding positional encoding, the model employs a single-layer MHA mechanism. The output from the MHA layer is then subjected to mean pooling to derive a global representation of the state. This global representation is then processed similarly to our primary model, passing through a feed-forward layer and a sigmoid function to predict the probability. The computation is as follows:
\begin{equation}
    \overline{h}_{i, m} = h_{i, m} + \text{PE}(i),
\end{equation}
\begin{equation}
    \widetilde{h}_{i,m} = \text{MHA}_{i,m}\left( \overline{h}_{1, 1},...,\overline{h}_{n, \left| M_n \right|} \right),
\end{equation}
\begin{equation}
    \textit{probability} = \text{sigmoid}\left( \text{FF}\left( \frac{1}{\sum_{i=1}^{n} \left|M_{i}\right|} \sum_{i=1}^{n} \sum_{m=1}^{\left| M_{i}\right|}\widetilde{h}_{i,m} \right) \right),
\end{equation}
where PE is defined for each position $i$ in the sequence as a vector of dimension $d_{\text{model}}$, with each element given by:
\begin{equation}
\text{PE}(i)[2k] = \sin\left(\frac{i}{10000^{2k/d_{\text{model}}}}\right),
\end{equation}
\begin{equation}
\text{PE}(i)[2k+1] = \cos\left(\frac{i}{10000^{2k/d_{\text{model}}}}\right),
\end{equation}
for \(k = 0, 1, \ldots, \frac{d_{\text{model}}}{2}-1\). Here, \(i\) represents the position, and \(d_{\text{model}}\) is the dimension of the model. For the detailed implementation of positional encoding, please refer to \cite{vaswani2017attention}.
\end{document}